\documentclass[letterpaper]{article} 
\usepackage{aaai2026}  
\usepackage{times}
\usepackage{helvet}
\usepackage{courier}
\usepackage{xcolor}
\usepackage{times}  
\usepackage{helvet}  
\usepackage{courier}  
\usepackage[hyphens]{url}  
\usepackage{graphicx} 
\usepackage{natbib}  
\usepackage{caption} 
\usepackage{algorithm}
\usepackage{amsmath, amssymb}
\usepackage{booktabs}
\usepackage{multirow}
\usepackage{xcolor}
\usepackage{float}
\usepackage{afterpage}
\usepackage{algpseudocode}
\newtheorem{lemma}{Lemma}
\usepackage{newfloat}
\usepackage{listings}
\DeclareCaptionStyle{ruled}{labelfont=normalfont,labelsep=colon,strut=off} 
\floatstyle{ruled}
\newfloat{listing}{tb}{lst}{}
\floatname{listing}{Listing}
\title{RAGFort: Dual-Path Defense Against Proprietary Knowledge Base Extraction\\ in Retrieval-Augmented Generation}
\author {
    Qinfeng Li\textsuperscript{\rm 1}\equalcontrib,\hspace{0.4em}
    Miao Pan\textsuperscript{\rm 1}\equalcontrib,\hspace{0.4em}
    Ke Xiong\textsuperscript{\rm 1}\equalcontrib,\hspace{0.4em}
    Ge Su\textsuperscript{\rm 1},\\
    Zhiqiang Shen\textsuperscript{\rm 2},\hspace{0.4em}
    Yan Liu\textsuperscript{\rm 2},\hspace{0.4em}
    SUN Bing\textsuperscript{\rm 3},\hspace{0.4em}
    Hao Peng\textsuperscript{\rm 4},\hspace{0.4em}
    Xuhong Zhang\textsuperscript{\rm 1,5}\thanks{~Corresponding to \texttt{zhangxuhong@zju.edu.cn}}
}

\affiliations {
    \textsuperscript{\rm 1}Zhejiang University\\
    \textsuperscript{\rm 2}Ant Group\\
    \textsuperscript{\rm 3}Universal Identification Technology (Hangzhou) Co., Ltd.\\
    \textsuperscript{\rm 4}Zhejiang Normal University\\
    \textsuperscript{\rm 5}Ningbo Global Innovation Center, Zhejiang University\\
}

\usepackage{bibentry}

\begin{document}

\maketitle

\begin{abstract}
Retrieval-Augmented Generation (RAG) systems deployed over proprietary knowledge bases face growing threats from reconstruction attacks that aggregate model responses to replicate knowledge bases. Such attacks exploit both intra-class and inter-class paths—progressively extracting fine-grained knowledge within topics and diffusing it across semantically related ones, thereby enabling comprehensive extraction of the original knowledge base. However, existing defenses target only one path, leaving the other unprotected. We conduct a systematic exploration to assess the impact of protecting each path independently and find that joint protection is essential for effective defense. Based on this, we propose RAGFort, a structure-aware dual-module defense combining \textit{contrastive reindexing} for inter-class isolation and \textit{constrained cascade generation} for intra-class protection. Experiments across security, performance, and robustness confirm that RAGFort significantly reduces reconstruction success while preserving answer quality, offering comprehensive defense against knowledge base extraction attacks.
\end{abstract}

\noindent\textbf{Code} — \url{https://github.com/happywinder/RAGFort}

\section{Introduction}

Large Language Models (LLMs) have demonstrated remarkable capabilities across a wide range of natural language processing tasks, particularly in open-domain question answering and knowledge reasoning~\cite{brown2020language,openai2023gpt4}. To enhance factual accuracy and controllability, Retrieval-Augmented Generation (RAG) has emerged as a widely adopted paradigm~\cite{lewis2020retrieval,izacard2021faithful}, enabling models to retrieve external knowledge to support answer generation. Beyond public knowledge like Wikipedia, RAG systems are increasingly deployed in high-value domains, such as healthcare~\cite{singhal2022medpalm}, and finance~\cite{wu2023bloomberggpt}, where the knowledge base is proprietary, domain-specific, and constitutes a core intellectual asset for commercial applications.

However, the privatized nature of these knowledge bases exposes a new vulnerability~\cite{ni2025towards}. Specifically, recent studies reveal that even by interacting with the RAG system in a black-box manner, attackers can extract a proxy knowledge base by aggregating responses from the RAG system~\cite{zeng2024good,cohen2024unleashing}. In particular, attackers may utilize the extracted knowledge base to construct functionally equivalent RAG systems~\cite{jiang2024rag}, thereby effectively replicating and misusing the original RAG system's capabilities. Such replication enables competitors to offer highly similar services, leading to direct financial losses and substantially eroding the competitive edge of the original provider. We refer to this threat as the \textit{knowledge base extraction attack}.

Specifically, recent studies show that attackers can extract information from a knowledge base by collecting and combining responses generated by the RAG system. For example, document extraction attacks have shown that attackers can craft queries that specifically target certain information. By prompting the model with these carefully designed queries, attackers can induce the RAG system to generate responses that contain or reveal underlying knowledge base content~\cite{zeng2024good}. 
More alarmingly, some emerging methods pose a more systemic extraction of the entire knowledge base rather than focusing on only isolated pieces of content. To achieve this, these attacks not only iteratively refine queries to extract detailed information within a specific targeted topic (referred to as  \textbf{\textit{intra-class extraction}}), but also gradually expand their query scope to cover other semantically related topics. Specifically, attackers leverage previously extracted knowledge to recursively generate queries targeting new semantically related topics. By iteratively expanding the extraction boundary, the attack gradually propagates through the knowledge base, eventually achieving nearly complete coverage (referred to as \textbf{\textit{inter-class extraction}}).
For example, RAG-Thief~\cite{jiang2024rag} employs an automated agent that starts with an initial adversarial query, stores the extracted chunks in memory, and then analyzes these chunks to generate more targeted queries for uncovering missing or previously unknown information. By iteratively updating its memory, the agent can both fill in missing details within the same topic (i.e., intra-class extraction) and gradually expand extraction to neighboring topics that were previously unknown or unqueried (i.e., inter-class extraction), ultimately reconstructing nearly the entire knowledge base.

However, existing defense strategies are still at a preliminary stage, typically focusing on individual attack paths rather than providing comprehensive protection. For example, intra-class defenses, such as paraphrasing or summarizing retrieved content, reduce leakage within individual chunks but fail to prevent cross-topic aggregation~\cite{zeng2024good,zeng2024mitigating}. Conversely, inter-class defenses restrict retrieval to closely related content, limiting topic drift but allowing in-depth extraction within a single class~\cite{zeng2024good}, such as limiting retrieval to the few most relevant chunks. 
Worse still, existing intra-class and inter-class defense schemes are difficult to combine effectively. For example, \textit{Set Distance Threshold}~\cite{zeng2024good}, as an inter-class protection, relies on strictly limiting the retrieval distance so that only highly relevant chunks are returned, which essentially forces the RAG system to \textbf{focus on only a few chunks} and prevents extraction from expanding to other topics. In contrast, to achieve intra-class protection, \textit{Summarization with Relevant Query}~\cite{zeng2024good} emphasizes generating diverse or aggregated content across multiple chunks within the same topic (i.e., through abstractive summarization) to prevent attackers from reconstructing the precise information of any single chunk, and thus inherently requires the RAG system \textbf{not to focus on only a few chunks}. As a result, this fundamental conflict poses significant challenges for combining intra-class and inter-class defenses in practice. Nevertheless, no prior work has systematically compared the two defense paths or explored integrated strategies that jointly secure both protection paths.

To bridge this gap, we first pose a key question: \textit{Between intra-class and inter-class protections, which is more critical for safeguarding proprietary knowledge base in RAG systems, and can integrated protection provide stronger security than either path alone?} To explore this, we design a targeted experiment that masks portions of knowledge along each attack path, and assess the resulting impact on answer quality.
Our findings reveal a key insight: protecting only one path offers limited resistance, whereas joint protection along both paths reduces the effectiveness of the proxy knowledge base. These results indicate that intra-class robustness and inter-class isolation are complementary, and that effective defense requires their coordination.

Based on this insight, we propose a structure-aware dual-module defense framework, RAGFort, as shown in Figure~\ref{figure_method} (see also Appendix~\ref{appendix:case_study} for an illustrative example).
For inter-class protection, we introduce a structure-aware reindexing strategy that reorganizes the retriever’s dense index to enhance semantic separability between topic classes. This transformation aims to reinforce topic boundaries, making it harder for attackers to expand queries from one class to another, thereby preventing inter-class extraction attacks. Specifically, the method begins with unsupervised HDBSCAN clustering to identify latent topic structures, followed by supervised contrastive learning to train an embedding encoder that preserves inter-category margins. 

For intra-class protection, we propose a cascaded generation framework that combines a lightweight draft model with a robust reference model. In this framework, the draft model first proposes candidate tokens, while the reference model selectively filters or replaces high-risk tokens. Specifically, when the draft model exhibits excessive uncertainty or its output contains sensitive content, the reference model intervenes and generates safer tokens. Essentially, we have designed a near-optimal and efficient rejection strategy that fuses the output distributions of the draft model and the reference model. The new distribution significantly reduces the probability of sensitive content being output and improves the model's inference performance.

We validate the RAGFort through comprehensive experiments on security, system performance, and ablation studies. For security, we evaluate RAGFort under advanced black-box extraction attacks and show that it reduces the proportion of knowledge base entries recovered by attackers by more than half compared to prior defenses. For system performance, RAGFort consistently preserves high answer quality while also maintaining generation efficiency, thus ensuring practical usability. 
Through extensive ablation, we show that each module contributes to knowledge protection, and their combination delivers strong, complementary effects. 
Overall, our contributions are summarized as follows:

\begin{itemize}

\item To the best of our knowledge, this is the first systematic study that defends against knowledge base extraction in RAG systems. This overlooked vulnerability poses significant threats in real-world deployments and leads to significant commercial risks for RAG providers.

\item We identify and formalize two critical paths to protect the RAG knowledge base: inter-class isolation and intra-class protection. We design a targeted experiment to validate the role of each path and demonstrate that their combined defense is essential for protecting knowledge base extraction attacks.

\item We propose a structure-aware dual-module defense that jointly addresses inter-class and intra-class threats. It combines contrastive reindexing to enhance topic-level isolation and a cascade framework to suppress sensitive content generation under adversarial prompts.

\item We conduct a comprehensive empirical evaluation of RAGFort. Our evaluation demonstrates that RAGFort significantly enhances knowledge base security while preserving RAG system performance. Ablation studies confirm that each module is effective on its own and further enhances overall protection when combined.
\end{itemize}



\section{Preliminaries}

\begin{figure*}[ht]
    \centering
    \includegraphics[width=1\textwidth]{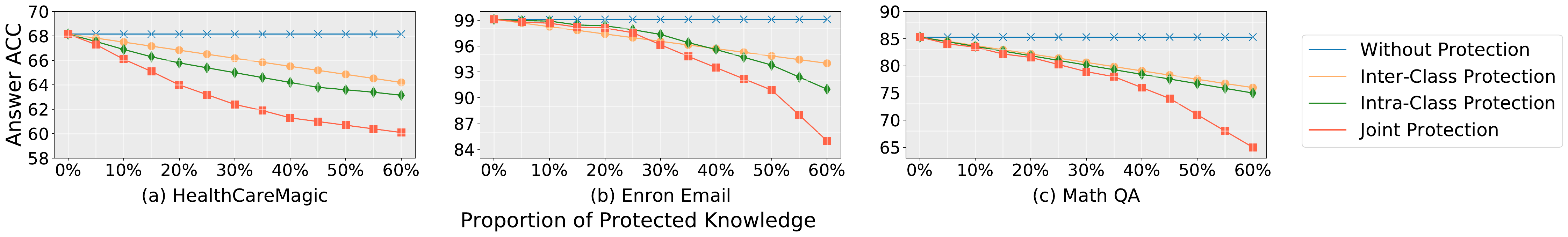}
    \caption{Dual-Path Protection Diagram. This diagram illustrates the structure-aware dual-module defense framework (RAGFort) for protecting data across vertical and horizontal axes.}
    \label{fig:dual-path-protection}
\end{figure*}

\subsubsection{Knowledge Base Extraction Attack.}
RAG systems generate answers by first retrieving relevant chunks from an external knowledge base via a retriever module and then using a language model to compose the final response.
While RAG systems enhance factual generation by retrieving external knowledge, they also expose proprietary content to extraction attacks. Specifically, recent studies show that adversaries can reconstruct knowledge bases by collecting and combining RAG responses to carefully crafted or paraphrased queries. By systematically piecing together related fragments returned by the system, attackers are able to recover large portions of the original content~\cite{ni2025towards,zeng2024good,jiang2024rag}. Emerging methods further streamline this process by using adaptive querying agents that automatically generate and refine queries to explore both within individual topics (intra-class) and across different topics (inter-class)~\cite{jiang2024rag,cohen2024unleashing}. This two-path strategy significantly improves the fidelity and coverage of the extraction.

\subsubsection{Existing Defense.}
Existing defense strategies remain at an early stage, limited to a single path, with few truly systematic approaches available. \emph{Intra-class} methods paraphrase or summarize individual chunks to obscure fine-grained content, or replace them altogether with synthetic data~\cite{zeng2024good,zeng2024mitigating}, while \emph{inter-class} defenses restrict retrieval to highly relevant chunks via distance thresholds, implicitly masking broader topical structures~\cite{zeng2024good}. However, current approaches address only one direction of attack in isolation. No unified framework exists to jointly defend across both axes, leaving RAG systems vulnerable to coordinated extraction strategies.

\subsubsection{Cascade Generation.}
For intra-class protection strategies, we adopt a cascade generation mechanism, namely a two-stage generation strategy~\cite{narasimhan2024faster}, to control the degree of knowledge exposure during the generation process. In the cascade framework, a lightweight draft model first proposes a set of candidate tokens, which are then selectively validated or rejected by a more powerful verifier model. If the verifier model chooses to reject, it generates a higher-quality token. This approach achieves faster generation speeds by enabling parallel evaluation while maintaining output quality. Unlike previous approaches that emphasized the efficiency of inference and quality cascading, we primarily modified the rejection rules in the cascading system to reduce the probability of retrieving sensitive content from the knowledge base during generation.

\subsubsection{Threat Model.}
In this paper, we consider two parties: the defender and the attacker.
Unlike privacy attacks that target individual records, we consider attackers whose objective is to extract the entire underlying knowledge base of the RAG system. The attacker is restricted to a black-box setting and can only access the RAG system via API queries, which aligns with real-world deployments of commercial and enterprise RAG services.
Defenders are system providers with full access to the RAG pipeline (including retrievers, databases, and language models). The defense objective is to protect proprietary knowledge bases from reconstruction while ensuring usability for legitimate users.



\section{Comparison of Different Defense Paths}


To investigate which protection, intra-class or inter-class, is more critical for safeguarding proprietary knowledge in RAG systems, and whether integrated protection offers stronger security than either path alone, we design an experiment that evaluates the impact of different protection paths by selectively masking portions of the knowledge base and analyzing the residual utility of protected knowledge bases.

\subsection{Experimental Design}
We begin with a complete knowledge base $\mathcal{K}_{\text{full}}$ constructed from domain-specific datasets. 
From this base, we derive three partially leaked variants by systematically removing knowledge portions considered well-protected, which are thus excluded from the proxy knowledge bases. The remaining parts constitute the knowledge bases that the attacker has successfully reconstructed through model stealing.

\begin{itemize}
\item \textbf{Without Protection}: No entries are removed or masked; the attacker obtains the full original knowledge base without any restrictions.
\item \textbf{Inter-class Protection}: Removes detailed entries within each category (e.g., specific conditions under "cardiology"), simulating intra-class protection.
\item \textbf{Intra-class Protection}: Removes entire topic categories (e.g., "dermatology" or "algebra"), simulating inter-class protection.
\item \textbf{Joint Protection}: Applies both inter-class and intra-class masking to simulate two-path coordinated defense.
\end{itemize}

Each variant is used to build an independent RAG system, and we evaluate its ability to correctly answer questions. \textit{Higher Answer ACC} indicates that the proxy knowledge base still retains substantial utility, suggesting that the corresponding protection strategy is \textit{less effective}. The detailed experimental setup is described in Section~\ref{Evaluation Setting}.

\subsection{Results and Analysis}

Figure~\ref{fig:dual-path-protection} presents the accuracy of attacker-constructed RAG systems under each protection path across three representative domains. The results reveal a clear trend: either inter-class or intra-class protection alone is insufficient to significantly limit knowledge base leakage. By contrast, the joint strategy substantially reduces the utility of the stolen knowledge base.
Specifically, the performance of the proxy knowledge base declines as the proportion of protected knowledge increases, and this decline is much more pronounced with joint protection than with either single-path strategy. This difference grows even further as more data is protected.
These findings empirically confirm our central hypothesis: \textbf{\textit{inter-class and intra-class protection are complementary, and only by combining both protection paths can protection achieve comprehensive and effective defense.}}

\begin{figure*}[t]
\centering
\includegraphics[width=\textwidth]
{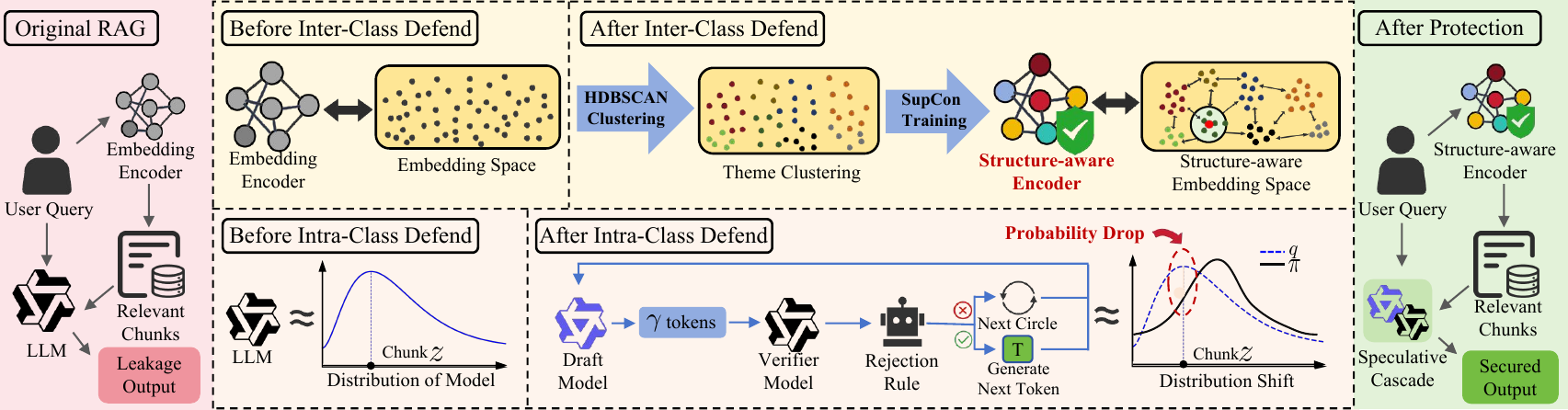} 

\caption{An overview of RAGFort. For inter-class protection, we introduce a structure-aware encoder trained to cluster and separate topics in the embedding space, making it harder for attackers to retrieve content across categories. For intra-class protection, candidate tokens proposed by a draft model are rigorously screened by a verifier model through a tailored rejection rule, which blocks sensitive or risky outputs.}
\label{figure_method}
\end{figure*}

\section{Our Design: RAGFort}

In this section, we introduce RAGFort, a structure-aware dual-path defense framework to counter knowledge base reconstruction attacks in RAG systems. 

\subsection{Overview}
RAGFort is achieved by integrating a contrastive reindexing mechanism with a constrained generation cascade, each targeting a different path of the knowledge base, as shown in Figure \ref{figure_method}. An additional example is provided in Appendix~\ref{appendix:case_study}.
Specifically, for \textit{inter-class protection}, we restructure the retriever’s index to enforce stronger semantic separation across topic categories. This process begins with unsupervised clustering using HDBSCAN to reveal latent topic structures. Then, we train a supervised contrastive encoder that preserves inter-cluster margins, enabling better class discrimination in the embedding space. As a result, it becomes significantly harder for attackers to retrieve content from unrelated topics.

For \textit{intra-class protection}, we develop a cascade generation mechanism that improves both generation quality and defense robustness. This design combines a fast draft model with a stronger verifier model. During inference, the draft model proposes candidate tokens, which are selectively validated by the reference model based on a rejection rule. To mitigate the generation of sensitive content in cascaded systems, we propose a computationally efficient rejection rule and provide theoretical guarantees that its performance closely approximates that of the optimal rule.

\subsection{Inter-Class Protection}
Our inter-class protection method proceeds in three stages: unsupervised clustering, supervised contrastive learning, and index replacement.

\subsubsection{Pseudo-Labeling via HDBSCAN Clustering.}

We begin by applying unsupervised clustering to identify latent topic categories within the knowledge base. First, each chunk $k_i \in \mathcal{K} = {k_1, \ldots, k_N}$ is embedded using a pretrained encoder (e.g., Sentence-BERT) to obtain dense representations $E(k_i)$. Next, we use Hierarchical Density-Based Spatial Clustering of Applications with Noise (HDBSCAN) \cite{campello2013density} to partition the chunks into $C$ clusters ${\mathcal{C}_1, ..., \mathcal{C}_C}$, thereby assigning a pseudo-label $y_i \in {1,...,C}$ to each $k_i$. Notably, HDBSCAN is chosen due to its capability to automatically infer the number of clusters, handle outliers, and adapt to high-dimensional semantic spaces.

\subsubsection{Learning Class-Separable Embeddings via SupCon Loss.}
Using the pseudo-labels from clustering, we train a new encoder with supervised contrastive learning (SupCon) to produce class-discriminative embeddings. Given a mini-batch of encoded chunks $\{z_1, \ldots, z_B\}$ with corresponding cluster labels $\{y_1, \ldots, y_B\}$, the SupCon loss is defined as:
{\footnotesize
\begin{equation}
\mathcal{L}_{\text{sup}} = \sum_{i=1}^{B} -\frac{1}{|P(i)|} \sum_{p \in P(i)} \log \frac{\exp(\text{sim}(z_i, z_p)/\tau)}{\sum_{a \in A(i)} \exp(\text{sim}(z_i, z_a)/\tau)},
\end{equation}
}
where $P(i)$ is the set of samples with the same label as $i$ (excluding $i$), $A(i)$ includes all samples in the batch except $i$, and $\text{sim}(\cdot,\cdot)$ denotes cosine similarity. $\tau$ is the temperature scaling factor. This contrastive objective strengthens the separation between different semantic categories, improving the structural integrity of indexed representations.

\subsubsection{Reindexing with Tokenizer-Aligned Contrastive Embeddings.}

After contrastive training, we obtain a structure-aware encoder $f_{\text{sup}}(\cdot)$ that outputs enhanced chunk embeddings $z_i' = f_{\text{sup}}(k_i)$. We then construct a new dense index for the retriever using these updated representations. Notably, these embeddings are generated based on a custom tokenizer aligned with the SupCon training process.

To preserve the generative performance of the original RAG system, we do not modify the query encoder or the decoder input pipeline. Instead, during inference, the retriever operates on the SupCon-based index, while the generator still consumes the original (non-contrastive) chunk content encoded using the original tokenizer.

\subsection{Intra-Class Protection via Cascade}

To enhance intra-class protection in RAG systems, we introduce a cascaded generation framework combining a fast draft model with a robust reference model. Let $q(z|x_{<t},z)$ denote the probability of generating chunk $z$ under the original draft model $q$. The objective of cascade is to construct a new generation distribution $\pi=(1 - r(x_{<t},z)) \cdot q_t + r(x_{<t},z) \cdot p_t$, where $r(x_{<t},z)$ is a rejection rule that can only take the values 0 or 1. If we want to improve inference efficiency and generation quality and reduce the sampling probability of $\pi(z|x_{<t},z)$, we need to design an optimal rejection rule $r$. The optimal rejection rule $r$ should minimize the loss of the large model supervision under the cascade distribution $\pi$. To control the cost of fallback to $p$, we minimize the expected loss under a constraint on the rejection budget $B$. Meanwhile, to reduce the probability of chunk $z$ being generated, we introduce the threshold $C$, which requires that the probability of draft model generating sensitive content be significantly lower than that of reference model:
\begin{equation}\label{eq:constrained}
\begin{aligned}
\min_r \quad & \mathbb{E}_{y \sim \mathbb{P}(\cdot \mid x_{<t},z)}[
(1 - r(x_{<t},z)) \cdot \ell(y, q_t) \\&+ r(x_{<t},z) \cdot \ell(y, p_t)] \\
\text{s.t.} \quad & r(x_{<t},z) \cdot D_{\text{TV}}(p_t, q_t) \leq B,\\
& (1-r(x_{<t},z)) \cdot \frac{q_t(z)}{p_t(z)}\leq C.
\end{aligned}
\end{equation}
\begin{algorithm}[t]
\caption{\textsc{Cascades}}\label{Cascades}
\textbf{Input:} Draft model $q$, Reference model $p$, Rejection rule $r$, Chunk $z$, Prefix $x_{<t}$, Block size $\gamma$ \\
\textbf{Output:} $x_t, \ldots, x_{t + j^*}$
\begin{algorithmic}[1]
\For{$j = 0$ to $\gamma - 1$}
    \State Compute $q_{t+j}(\cdot \mid x_{<t+j}, z)$ and $q_t(z)$
\EndFor
\State Compute $p_t(z)$
\For{$j = 1$ to $\gamma$}
    \State Compute $p_{t+j}(\cdot \mid x_{<t+j}, z)$ and $r_{t+j}(x_{<t+j},z)$
    \State $\pi_{t+j}(\cdot) \gets (1-r(x_{<t+j}, z))\cdot q+r(x_{<t+j}, z)\cdot p$
\EndFor
\State $j^*=\gamma$
\For{$j = 0$ to $\gamma-1$}
    \If{$r_{t+j}(x_{<t+j},z)=0$}
    \State \textbf{continue}
    \Else
    \State $a_j \sim \text{Bernoulli} \left( \min \left\{ 1, \frac{p_{t+j}(x_{t+j})}{q_{t+j}(x_{t+j})} \right\} \right)$
    \If{$a_j=0$}
        $j^*=j$ and break
    \EndIf
    \EndIf
\EndFor
\If{$j^* < \gamma$}
    \State $p_{\text{cas}}(\cdot) \gets \text{norm}\left( \max\{ 0, \pi_{t+j^*}(\cdot) - q_{t+j^*}(\cdot) \} \right)$
\Else
    \State $p_{\text{cas}}(\cdot) \gets \pi_{t+\gamma}$
\EndIf
\State Sample $x_{t+j^*} \sim p_{\text{ref}}(\cdot)$
\end{algorithmic}
\end{algorithm}
\subsubsection{Analysis of Rejection Rule.}
To ease optimization, we transform it into an unconstrained problem using Lagrangian relaxation:
\begin{equation}\label{eq:lagrangian}
{\footnotesize
\begin{aligned}
&L(r; x_{<t},z) \\=& \mathbb{E}_{y \sim \mathbb{P}(\cdot \mid x_{<t},z)} [r(x_{<t},z) \cdot (\ell(y, p_t) + \alpha \cdot D_{\text{TV}}(p_t, q_t)) \\+& (1 - r(x_{<t},z)) \cdot (\ell(y, q_t)+\eta\cdot\frac{q_t(z)}{p_t(z)})],
\end{aligned}
}
\end{equation}
where $\alpha$ and $\eta$ are Lagrangian multipliers. The optimal $r$ for minimizing~\eqref{eq:lagrangian} is given by:
\begin{lemma}\label{optimal}
Let $\mathbb{P}_t(y) = \mathbb{P}(y \mid x_{<t}, z)$ is real distribution. Then the optimal rejection rule is:
\begin{equation}
\begin{aligned}
&r^*(x_{<t}, z)\\ =& 
\begin{cases}
1 & \text{if } \mathbb{E}_{y \sim \mathbb{P}_t} [\ell(y, q_t)] > \mathbb{E}_{y \sim \mathbb{P}_t} [\ell(y, p_t)]\\ &+ \alpha \cdot D_{\text{TV}}(p_t, q_t)-\eta\cdot \frac{q_t(z)}{p_t(z)}, \\
0 & \text{otherwise.}
\end{cases}
\end{aligned}
\end{equation}
\end{lemma}
Intuitively, the cascade system will only choose to generate a verifier model when the expected loss of the verifier model is significantly too high. Since $\mathbb{P}_t$ is unknown, we approximate the expected loss using confidence scores:
\begin{equation}\label{eq:plug-in-rule}
{\footnotesize 
\begin{aligned}
&\hat{r}(x_{<t},z) = 1 \quad \textbf{if and only if} \\
\max_{y} &q_t(y) < \max_{y} p_t(y) - \alpha \cdot D_{\text{TV}}(p_t, q_t)+\eta\cdot \frac{q_t(z)}{p_t(z)}.
\end{aligned}
}
\end{equation}
This rule triggers fallback when $p_t$ is sufficiently more confident than $q_t$. Next, we analyze the regret (performance gap) between the simple rejection rule $\hat{r}$ and the optimal rejection rule $r$.
\begin{lemma}\label{bound}
Let $\mathbb{P}_t(y) = \mathbb{P}(y \mid x_{<t}, z)$ is real distribution. Then for fixed $x_{<t}$ and $z$, we have
\begin{equation}
\begin{aligned}
&L(\hat{r}; x_{<t}, z) 
- \min_r L(r; x_{<t}, z) 
 \\ \leq &\max_{y \in \mathcal{Y}} |\mathbb{P}_t(y) - q_t(y)| 
+ \max_{y \in \mathcal{Y}} |\mathbb{P}_t(y) - p_t(y)|.
\end{aligned}
\end{equation}
\end{lemma}
This result shows that the regret is small when both $q_t$ and $p_t$ are close to the ground-truth distribution $\mathbb{P}_t$. 

\begin{table*}[ht]
\renewcommand{\arraystretch}{0.73}
\centering
\resizebox{\linewidth}{!}{
\begin{tabular}{lcccccccccccccccc}
\toprule
\multirow{2}{*}{\parbox{1cm}{Datasets}} & \multirow{2}{*}{\parbox{1cm}{Models}} & \multicolumn{2}{c}{\textbf{Without protection}} & & \multicolumn{2}{c}{\textbf{RAGFort (Ours)}} & & \multicolumn{2}{c}{\textbf{Re-ranking}} & & \multicolumn{2}{c}{\textbf{Summarization}} \\ 
\cline{3-4} \cline{6-7} \cline{9-10} \cline{12-13} \noalign{\smallskip} & & Worm-attack & RAG-Thief & & Worm-attack & RAG-Thief & & Worm-attack & RAG-Thief & & Worm-attack & RAG-Thief \\
\cline{1-13} \noalign{\smallskip}
\multirow{2}{*}{HealthCareMagic} 
& Qwen-14B & 17.60\scriptsize\textcolor{gray!150}{$\pm$1.58} & 57.16\scriptsize\textcolor{gray!150}{$\pm$2.13} & & 8.84\scriptsize\textcolor{gray!150}{$\pm$0.91} & 27.96\scriptsize\textcolor{gray!150}{$\pm$1.32} & & 16.28\scriptsize\textcolor{gray!150}{$\pm$1.17} & 56.44\scriptsize\textcolor{gray!150}{$\pm$2.08} & & 15.44\scriptsize\textcolor{gray!150}{$\pm$1.20} & 47.24\scriptsize\textcolor{gray!150}{$\pm$1.81} \\
& DeepSeek-R1-8B & 13.92\scriptsize\textcolor{gray!150}{$\pm$1.29} & 51.64\scriptsize\textcolor{gray!150}{$\pm$1.92} & & 8.56\scriptsize\textcolor{gray!150}{$\pm$0.83} & 26.72\scriptsize\textcolor{gray!150}{$\pm$1.18} & & 12.64\scriptsize\textcolor{gray!150}{$\pm$0.94} & 49.28\scriptsize\textcolor{gray!150}{$\pm$1.76} & & 12.88\scriptsize\textcolor{gray!150}{$\pm$0.91} & 42.52\scriptsize\textcolor{gray!150}{$\pm$1.54} \\
& Gemma-3-27B & 18.36\scriptsize\textcolor{gray!150}{$\pm$1.33} & 59.88\scriptsize\textcolor{gray!150}{$\pm$1.71} & & 9.76\scriptsize\textcolor{gray!150}{$\pm$0.88} & 29.44\scriptsize\textcolor{gray!150}{$\pm$1.11} & & 18.24\scriptsize\textcolor{gray!150}{$\pm$0.96} & 57.36\scriptsize\textcolor{gray!150}{$\pm$1.84} & & 16.12\scriptsize\textcolor{gray!150}{$\pm$0.97} & 47.28\scriptsize\textcolor{gray!150}{$\pm$1.62} \\
\cline{1-13} \noalign{\smallskip}
\multirow{2}{*}{Enron Email} 
& Qwen-14B & 31.60\scriptsize\textcolor{gray!150}{$\pm$1.73} & 52.60\scriptsize\textcolor{gray!150}{$\pm$1.95} & & 14.30\scriptsize\textcolor{gray!150}{$\pm$1.01} & 20.60\scriptsize\textcolor{gray!150}{$\pm$0.89} & & 28.65\scriptsize\textcolor{gray!150}{$\pm$1.58} & 44.35\scriptsize\textcolor{gray!150}{$\pm$1.49} & & 25.35\scriptsize\textcolor{gray!150}{$\pm$1.43} & 46.75\scriptsize\textcolor{gray!150}{$\pm$1.66} \\
& DeepSeek-R1-8B & 28.75\scriptsize\textcolor{gray!150}{$\pm$1.45} & 47.85\scriptsize\textcolor{gray!150}{$\pm$1.88} & & 10.15\scriptsize\textcolor{gray!150}{$\pm$0.75} & 17.65\scriptsize\textcolor{gray!150}{$\pm$0.69} & & 23.45\scriptsize\textcolor{gray!150}{$\pm$1.12} & 43.75\scriptsize\textcolor{gray!150}{$\pm$1.37} & & 24.75\scriptsize\textcolor{gray!150}{$\pm$1.16} & 40.65\scriptsize\textcolor{gray!150}{$\pm$1.31} \\
& Gemma-3-27B & 32.35\scriptsize\textcolor{gray!150}{$\pm$1.41} & 54.15\scriptsize\textcolor{gray!150}{$\pm$1.99} & & 14.40\scriptsize\textcolor{gray!150}{$\pm$0.71} & 24.55\scriptsize\textcolor{gray!150}{$\pm$0.76} & & 28.85\scriptsize\textcolor{gray!150}{$\pm$1.28} & 46.50\scriptsize\textcolor{gray!150}{$\pm$1.52} & & 27.60\scriptsize\textcolor{gray!150}{$\pm$1.13} & 47.30\scriptsize\textcolor{gray!150}{$\pm$1.43} \\
\cline{1-13} \noalign{\smallskip}
\multirow{2}{*}{Math QA} 
& Qwen-14B & 47.84\scriptsize\textcolor{gray!150}{$\pm$2.24} & 70.31\scriptsize\textcolor{gray!150}{$\pm$2.87} & & 29.06\scriptsize\textcolor{gray!150}{$\pm$1.02} & 38.13\scriptsize\textcolor{gray!150}{$\pm$1.41} & & 39.49\scriptsize\textcolor{gray!150}{$\pm$1.90} & 74.69\scriptsize\textcolor{gray!150}{$\pm$2.99} & & 35.31\scriptsize\textcolor{gray!150}{$\pm$1.78} & 68.44\scriptsize\textcolor{gray!150}{$\pm$2.67} \\
& DeepSeek-R1-8B & 43.13\scriptsize\textcolor{gray!150}{$\pm$2.11} & 67.81\scriptsize\textcolor{gray!150}{$\pm$2.54} & & 28.44\scriptsize\textcolor{gray!150}{$\pm$1.39} & 37.81\scriptsize\textcolor{gray!150}{$\pm$1.87} & & 40.00\scriptsize\textcolor{gray!150}{$\pm$1.97} & 58.75\scriptsize\textcolor{gray!150}{$\pm$2.01} & & 41.88\scriptsize\textcolor{gray!150}{$\pm$1.91} & 55.63\scriptsize\textcolor{gray!150}{$\pm$1.88} \\
& Gemma-3-27B & 48.75\scriptsize\textcolor{gray!150}{$\pm$2.26} & 69.06\scriptsize\textcolor{gray!150}{$\pm$2.62} & & 29.69\scriptsize\textcolor{gray!150}{$\pm$0.98} & 38.44\scriptsize\textcolor{gray!150}{$\pm$1.38} & & 43.13\scriptsize\textcolor{gray!150}{$\pm$1.78} & 63.44\scriptsize\textcolor{gray!150}{$\pm$1.95} & & 43.44\scriptsize\textcolor{gray!150}{$\pm$1.51} & 64.69\scriptsize\textcolor{gray!150}{$\pm$1.57} \\
\cline{1-13} \noalign{\smallskip}
\multicolumn{2}{c}{\textbf{Relative Mean CRR}} & \multicolumn{2}{c}{\textbf{1}$\times$} &  & \multicolumn{2}{c}{\textbf{0.51}$\times$} &  & \multicolumn{2}{c}{\textbf{0.91}$\times$} &  & \multicolumn{2}{c}{\textbf{0.87}$\times$} \\
\bottomrule
\end{tabular}
}
\caption{Security assessment of RAGFort in preventing knowledge base extraction attack. We report the Chunk Recovery Rate (CRR, \%) of RAG systems under two black-box attack strategies: \textit{Worm-Attack} and \textit{RAG-Thief}. Each column block corresponds to a specific defense method—\textit{Without protection}, \textit{RAGFort (Ours)}, \textit{Re-ranking}, and \textit{Summarization}—while each sub-column reports CRR under a specific attack. Lower CRR $\downarrow$ indicates stronger protection.
The last row shows the average attack CRR as a multiple of the baseline without protection ($\times$).}
\label{security}
\end{table*}
The algorithm~\ref{Cascades} describes a cascading generation process in which the draft model first proposes a block of $\gamma$ tokens and calculates the probability $q_t$. For each token position, the verifier model concurrently computes the probability $p_t$ and the rejection rule $r$, and then uses $r$ to generate $\pi$. Subsequently, the earliest rejection position $j^*$ is sought. If a rejection occurs, a residual distribution is constructed to optimize token generation at $j^*$; otherwise, the draft output is retained, and the verifier model is used to generate the starting token for the next block. Importantly, we utilize the Bernoulli distribution during the rejection process. This ensures that the entire generation process starts from $q$ but ultimately generates tokens whose distribution does not deviate from the verifier model $p$~\cite{yan2024decoding, zhou2023distillspec}.

\section{Experiments}

\subsection{Evaluation Settings}
\label{Evaluation Setting}

\subsubsection{RAG Architecture and Implementation Details.}
We implement a standard RAG pipeline using the LangChain framework~\cite{chase2022langchain}. Each data sample (either a QA pair or a document) is treated as a distinct chunk in the knowledge base. At inference time, the built-in LangChain retriever, which operates on dense vector similarity, retrieves the top 5 relevant chunks for each query. The retrieved chunks are then concatenated with the question to form the input to the generator, following the prompt format: “\texttt{Context: [...] Question: [...] Answer:}”. For the generator, we evaluate three large language models: Qwen-14B~\cite{qwen2023}, a high-capacity open-source model; DeepSeek-R1-Distill-Llama-8B (abbreviated as \textit{DeepSeek-R1-8B})~\cite{guo2025deepseek}, a lightweight distilled variant of the LLaMA series optimized for chat-based generation; and Gemma-3-27B~\cite{team2025gemma}, a strong 27B-parameter model featuring enhanced reasoning and factual consistency through large-scale multi-domain instruction tuning. All RAG systems share the same retriever, prompting format, and decoding settings to ensure a fair comparison.
To implement cascade generation in RAGFort, we adopt a two-stage setup with a draft and a verifier model. For Qwen-14B, we use Qwen-7B as the draft model. For Gemma-3-27B, we use Gemma-3-4B as the draft model. For DeepSeek-R1-8B, the same model serves both roles, simulating scenarios where no smaller model is available. We apply different decoding temperatures for the draft and verifier. This setup shows that our defense remains effective even without auxiliary models.

\subsubsection{Dataset and Evaluation Protocol.}
We evaluate on three datasets spanning different domains:
(1) \textbf{HealthCareMagic}\cite{pampari2018emrqa}: A medical question-answering dataset containing symptom descriptions, diagnoses, and treatment suggestions across various specialties.
(2) \textbf{Enron Email}\cite{klimt2004enron}: A collection of corporate email communications covering diverse topics, organizational roles, and business contexts.
(3) \textbf{Math QA}~\cite{amini2019mathqa}: A dataset consisting of math word problems and multiple-choice solutions that require numerical reasoning and problem-solving skills.

Each RAG system is evaluated using QA pairs constructed from the corresponding dataset. We use GPT-4-turbo (\textit{gpt-4-0125-preview})~\cite{openai2024gpt4turbo} as an automatic evaluator to determine whether each generated answer is semantically consistent and factually aligned with the ground truth. A response is considered correct if it meets these criteria. The overall performance is measured by the percentage of correct responses, reported as \textbf{Answer Accuracy (ACC)}.

\subsubsection{Attack Strategies.}
We evaluate defense robustness under the two most widely recognized attack strategies:
(1) \textbf{Worm-Attack}\cite{cohen2024unleashing}: Utilizes adversarial, self-replicating prompts to trigger cascading extraction and propagation across RAG-based applications, simulating a computer worm in the GenAI ecosystem. Each experiment runs 1,000 attack rounds with 10 collision vector search iterations per round.
(2) \textbf{RAG-Thief}\cite{jiang2024rag}: An agent-based attack that maintains memory of extracted chunks and iteratively generates new adversarial queries through self-improvement, enabling scalable extraction from the knowledge base. Each round issues 10 queries, initialized with 300, 250, and 50 GPT-generated prompts for HealthCareMagic, EnronMail, and MathQA, respectively (final chunk sets: 2,500, 2,000, and 320).

\begin{table*}[t]
\renewcommand{\arraystretch}{0.73}
\centering
\resizebox{\linewidth}{!}{
\begin{tabular}{lcccccccccccccccc}
\toprule
\multirow{2}{*}{\parbox{1cm}{Datasets}} & \multirow{2}{*}{\parbox{1cm}{Models}} & \multicolumn{2}{c}{\textbf{Without protection}} & & \multicolumn{2}{c}{\textbf{RAGFort (Ours)}} & & \multicolumn{2}{c}{\textbf{RAGFort}\textsubscript{InterOnly}} & & \multicolumn{2}{c}{\textbf{RAGFort}\textsubscript{IntraOnly}} \\ 
\cline{3-4} \cline{6-7} \cline{9-10} \cline{12-13} \noalign{\smallskip} & & Worm-attack & RAG-Thief & & Worm-attack & RAG-Thief & & Worm-attack & RAG-Thief & & Worm-attack & RAG-Thief \\
\cline{1-13} \noalign{\smallskip}
\multirow{2}{*}{HealthCareMagic} 
& Qwen-14B & 17.60\scriptsize\textcolor{gray!150}{$\pm$1.58} & 57.16\scriptsize\textcolor{gray!150}{$\pm$2.13} & & 8.84\scriptsize\textcolor{gray!150}{$\pm$0.91} & 27.96\scriptsize\textcolor{gray!150}{$\pm$1.32} & & 14.52\scriptsize\textcolor{gray!150}{$\pm$0.62} & 41.16\scriptsize\textcolor{gray!150}{$\pm$1.71} & & 13.62\scriptsize\textcolor{gray!150}{$\pm$0.83} & 51.28\scriptsize\textcolor{gray!150}{$\pm$2.57} \\
& DeepSeek-R1-8B & 13.92\scriptsize\textcolor{gray!150}{$\pm$1.29} & 51.64\scriptsize\textcolor{gray!150}{$\pm$1.92} & & 8.56\scriptsize\textcolor{gray!150}{$\pm$0.83} & 26.72\scriptsize\textcolor{gray!150}{$\pm$1.18} & & 9.04\scriptsize\textcolor{gray!150}{$\pm$0.45} & 39.56\scriptsize\textcolor{gray!150}{$\pm$2.22} & & 11.51\scriptsize\textcolor{gray!150}{$\pm$0.68} & 42.36\scriptsize\textcolor{gray!150}{$\pm$2.67} \\
& Gemma-3-27B & 18.36\scriptsize\textcolor{gray!150}{$\pm$1.33} & 59.88\scriptsize\textcolor{gray!150}{$\pm$1.71} & & 9.76\scriptsize\textcolor{gray!150}{$\pm$0.88} & 29.44\scriptsize\textcolor{gray!150}{$\pm$1.11} & & 15.28\scriptsize\textcolor{gray!150}{$\pm$0.51} & 46.32\scriptsize\textcolor{gray!150}{$\pm$1.35} & & 14.72\scriptsize\textcolor{gray!150}{$\pm$0.73} & 51.56\scriptsize\textcolor{gray!150}{$\pm$2.31} \\
\cline{1-13} \noalign{\smallskip}
\multirow{2}{*}{Enron Email} 
& Qwen-14B & 31.60\scriptsize\textcolor{gray!150}{$\pm$1.73} & 52.60\scriptsize\textcolor{gray!150}{$\pm$1.95} & & 14.30\scriptsize\textcolor{gray!150}{$\pm$1.01} & 20.60\scriptsize\textcolor{gray!150}{$\pm$0.89} & & 19.75\scriptsize\textcolor{gray!150}{$\pm$0.90} & 33.25\scriptsize\textcolor{gray!150}{$\pm$1.25} & & 26.10\scriptsize\textcolor{gray!150}{$\pm$1.75} & 36.70\scriptsize\textcolor{gray!150}{$\pm$1.40} \\
& DeepSeek-R1-8B & 28.75\scriptsize\textcolor{gray!150}{$\pm$1.45} & 47.85\scriptsize\textcolor{gray!150}{$\pm$1.88} & & 10.15\scriptsize\textcolor{gray!150}{$\pm$0.75} & 17.65\scriptsize\textcolor{gray!150}{$\pm$0.69} & & 24.15\scriptsize\textcolor{gray!150}{$\pm$0.95} & 31.40\scriptsize\textcolor{gray!150}{$\pm$1.15} & & 22.55\scriptsize\textcolor{gray!150}{$\pm$1.65} & 27.85\scriptsize\textcolor{gray!150}{$\pm$2.60} \\
& Gemma-3-27B & 32.35\scriptsize\textcolor{gray!150}{$\pm$1.41} & 54.15\scriptsize\textcolor{gray!150}{$\pm$1.99} & & 14.40\scriptsize\textcolor{gray!150}{$\pm$0.71} & 24.55\scriptsize\textcolor{gray!150}{$\pm$0.76} & & 24.85\scriptsize\textcolor{gray!150}{$\pm$1.12} & 34.85\scriptsize\textcolor{gray!150}{$\pm$1.22} & & 27.35\scriptsize\textcolor{gray!150}{$\pm$1.30} & 45.05\scriptsize\textcolor{gray!150}{$\pm$1.95} \\
\cline{1-13} \noalign{\smallskip}
\multirow{2}{*}{Math QA} 
& Qwen-14B & 47.84\scriptsize\textcolor{gray!150}{$\pm$2.24} & 70.31\scriptsize\textcolor{gray!150}{$\pm$2.87} & & 29.06\scriptsize\textcolor{gray!150}{$\pm$1.02} & 38.13\scriptsize\textcolor{gray!150}{$\pm$1.41} & & 30.94\scriptsize\textcolor{gray!150}{$\pm$0.91} & 54.61\scriptsize\textcolor{gray!150}{$\pm$2.19} & & 43.75\scriptsize\textcolor{gray!150}{$\pm$1.51} & 65.63\scriptsize\textcolor{gray!150}{$\pm$2.58} \\
& DeepSeek-R1-8B & 43.13\scriptsize\textcolor{gray!150}{$\pm$2.11} & 67.81\scriptsize\textcolor{gray!150}{$\pm$2.54} & & 28.44\scriptsize\textcolor{gray!150}{$\pm$1.39} & 37.81\scriptsize\textcolor{gray!150}{$\pm$1.87} & & 41.25\scriptsize\textcolor{gray!150}{$\pm$1.61} & 52.19\scriptsize\textcolor{gray!150}{$\pm$1.68} & & 42.19\scriptsize\textcolor{gray!150}{$\pm$2.23} & 53.78\scriptsize\textcolor{gray!150}{$\pm$1.74} \\
& Gemma-3-27B & 48.75\scriptsize\textcolor{gray!150}{$\pm$2.26} & 69.06\scriptsize\textcolor{gray!150}{$\pm$2.62} & & 29.69\scriptsize\textcolor{gray!150}{$\pm$0.98} & 38.44\scriptsize\textcolor{gray!150}{$\pm$1.38} & & 37.81\scriptsize\textcolor{gray!150}{$\pm$1.43} & 54.06\scriptsize\textcolor{gray!150}{$\pm$2.02} & & 44.06\scriptsize\textcolor{gray!150}{$\pm$1.78} & 62.19\scriptsize\textcolor{gray!150}{$\pm$2.26} \\
\cline{1-13} \noalign{\smallskip}
\multicolumn{2}{c}{\textbf{Relative Mean CRR}} & \multicolumn{2}{c}{\textbf{1}$\times$} &  & \multicolumn{2}{c}{\textbf{0.51}$\times$} &  & \multicolumn{2}{c}{\textbf{0.75}$\times$} &  & \multicolumn{2}{c}{\textbf{0.83}$\times$} \\
\bottomrule
\end{tabular}
}
\caption{Ablation study of RAGFort defense modules against knowledge base extraction attacks.
We report the Chunk Recovery Rate (CRR, \%) for the full \textit{RAGFort} and its two path variants under different attack strategies, where a lower CRR indicates stronger protection.
\textit{RAGFort}\textsubscript{InterOnly} includes only the inter-class protection module (contrastive reindexing), while \textit{RAGFort}\textsubscript{IntraOnly} includes only the intra-class protection module (cascade).}
\label{Ablation Study}
\end{table*}

\begin{table}[t]
\renewcommand{\arraystretch}{0.71}
\centering
\setlength{\tabcolsep}{2pt}
\resizebox{\linewidth}{!}{
\begin{tabular}{lcccccc}
\toprule
\multirow{2}{*}{\parbox{1cm}{Datasets}} & \multirow{2}{*}{\parbox{1cm}{Models}} & \multicolumn{2}{c}{\textbf{Before protection}} & & \multicolumn{2}{c}{\textbf{After protection}} \\ 
\cline{3-4} \cline{6-7} \noalign{\smallskip}
& & ACC & FLOPs (T) & & ACC & FLOPs (T)\\
\cline{1-7} \noalign{\smallskip}
\multirow{2}{*}{HealthCareMagic} 
& Qwen-14B         & 
68.16\scriptsize\textcolor{gray!150}{$\pm$1.45} & 38.03\scriptsize\textcolor{gray!150}{$\pm$0.27} & & 66.57\scriptsize\textcolor{gray!150}{$\pm$1.22} & 19.40\scriptsize\textcolor{gray!150}{$\pm$0.35} \\
& DeepSeek-R1-8B   & 
61.36\scriptsize\textcolor{gray!150}{$\pm$0.61} & 20.37\scriptsize\textcolor{gray!150}{$\pm$0.23} & & 61.12\scriptsize\textcolor{gray!150}{$\pm$0.44} & 20.83\scriptsize\textcolor{gray!150}{$\pm$0.24} \\
& Gemma-3-27B   & 
69.64\scriptsize\textcolor{gray!150}{$\pm$0.75} & 84.83\scriptsize\textcolor{gray!150}{$\pm$0.58} & & 68.88\scriptsize\textcolor{gray!150}{$\pm$0.36} & 34.33\scriptsize\textcolor{gray!150}{$\pm$0.21} \\
\cline{1-7} \noalign{\smallskip}
\multirow{2}{*}{Enron Email} 
& Qwen-14B         & 
99.10\scriptsize\textcolor{gray!150}{$\pm$0.11} & 106.14\scriptsize\textcolor{gray!150}{$\pm$0.30} & & 98.85\scriptsize\textcolor{gray!150}{$\pm$0.25} & 56.84\scriptsize\textcolor{gray!150}{$\pm$0.40} \\
& DeepSeek-R1-8B   & 
97.50\scriptsize\textcolor{gray!150}{$\pm$0.15} & 56.90\scriptsize\textcolor{gray!150}{$\pm$0.15} & & 97.60\scriptsize\textcolor{gray!150}{$\pm$0.17} & 57.43\scriptsize\textcolor{gray!150}{$\pm$0.48} \\
& Gemma-3-27B   & 
99.25\scriptsize\textcolor{gray!150}{$\pm$0.21} & 236.47\scriptsize\textcolor{gray!150}{$\pm$0.86} & & 99.15\scriptsize\textcolor{gray!150}{$\pm$0.38} & 91.40\scriptsize\textcolor{gray!150}{$\pm$0.33} \\
\cline{1-7} \noalign{\smallskip}
\multirow{2}{*}{Math QA} 
& Qwen-14B         & 
85.31\scriptsize\textcolor{gray!150}{$\pm$1.63} & 47.09\scriptsize\textcolor{gray!150}{$\pm$0.35} & & 82.81\scriptsize\textcolor{gray!150}{$\pm$1.51} & 19.30\scriptsize\textcolor{gray!150}{$\pm$0.37} \\
& DeepSeek-R1-8B   & 
78.44\scriptsize\textcolor{gray!150}{$\pm$1.49} & 25.22\scriptsize\textcolor{gray!150}{$\pm$0.15} & & 77.19\scriptsize\textcolor{gray!150}{$\pm$2.47} & 25.37\scriptsize\textcolor{gray!150}{$\pm$0.36} \\
& Gemma-3-27B   & 
87.50\scriptsize\textcolor{gray!150}{$\pm$0.41} & 105.21\scriptsize\textcolor{gray!150}{$\pm$0.22} & & 85.94\scriptsize\textcolor{gray!150}{$\pm$1.26} & 31.95\scriptsize\textcolor{gray!150}{$\pm$0.31} \\
\cline{1-7} \noalign{\smallskip}
\end{tabular}

}
\caption{System performance before and after applying RAGFort protection.
We report the Answer Accuracy (ACC, \%) and computational cost (FLOPs (T)) of RAG systems with and without RAGFort.}
\label{Model Performance}
\end{table}

\subsubsection{Baseline Defenses.}

Our proposed method is compared to two recently proposed defense baselines: (1) \textbf{Re-ranking Protection}\cite{zeng2024good}: an inter-class protection that enforces semantic similarity thresholds during retrieval to filter loosely related documents and limit inter-class exposure. (2) \textbf{Summarization Protection}\cite{zeng2024mitigating}: Substitutes retrieved passages with abstract summaries to obscure fine-grained content and reduce intra-class leakage.

\subsubsection{Security Metrics.}

Following prior work~\cite{jiang2024rag}, we introduce the \textbf{Chunk Recovery Rate (CRR)}, which jointly captures both lexical and semantic similarity between generated outputs and original knowledge chunks. A chunk is counted as recovered only if it satisfies both of the following criteria:
(1) \textbf{ROUGE-L} (Lexical Similarity): The ROUGE-L score between the generated output and the original chunk exceeds a predefined threshold of 0.5.
(2) \textbf{Semantic Similarity}: The cosine similarity between the embedding vectors of the generated and original text, computed using a pre-trained model such as Sentence-BERT or BERTScore, exceeds a semantic threshold of 0.85.

\subsection{Security}
To evaluate defense effectiveness, we simulate knowledge extraction attacks and assess how well each method prevents unauthorized replication. The attacker interacts with the RAG system to reconstruct a proxy knowledge base, and we measure defense performance by the number of recovered chunks. Lower values indicate stronger protection.

As shown in Table~\ref{security}, RAGFort achieves the lowest Chunk Recovery Rate (CRR) across both Worm-Attack and RAG-Thief, demonstrating significantly stronger protection than existing defenses.
Specifically, RAGFort reduces the Relative Mean CRR to just 0.51× of the original. By comparison, re-ranking and summarization achieve only 0.91× and 0.87×, respectively, indicating substantially weaker protection. These results validate the effectiveness of our dual-path design: contrastive reindexing restricts inter-class exposure, while cascade suppresses intra-class leakage, jointly blocking both attack paths.

\subsection{System Performance}
To assess the impact on system performance of RAGFort, we compare QA accuracy and FLOPs before and after protection (as shown in Table~\ref{Model Performance}). Results show that RAGFort almost does not affect system performance: for all models and datasets, the ACC drops by less than 2 percentage points (e.g., Qwen-14B on HealthCareMagic: 68.16 → 66.57; DeepSeek-R1-8B: 61.36 → 61.12), and FLOPs remain nearly unchanged or even decrease.

We believe the small accuracy drop mainly arises from the gap between the draft and original models, not the defense itself. Specifically, Qwen-14B, which uses a smaller draft model, sees a slightly greater drop than DeepSeek-R1-8B, which uses the same model for both stages. Thus, we recommend stronger draft models when accuracy is a priority. Regarding efficiency, in the Qwen-14B group, the decrease in FLOPs results from replacing large-model inference with small-model inference, while in DeepSeek-R1-8B, which uses the same model as the draft model, FLOPs are nearly unchanged. This indicates that even when no smaller draft model is available, RAGFort introduces negligible overhead.

\subsection{Ablation Study}
To assess the individual contributions of each defense module, we perform an ablation study by isolating the inter-class and intra-class protection components. Similar to the security evaluation, we apply attacks on RAG systems equipped with only one defense module and measure the resulting chunk recovery rate and QA accuracy. This allows us to quantify the security gains provided by each component and validate their complementary roles.
As shown in Table~\ref{Ablation Study}, neither RAGFort\textsubscript{InterOnly} nor RAGFort\textsubscript{IntraOnly} alone provides sufficient defense. For instance, under RAG-Thief, the full RAGFort reduces CRR from 57.16\% to 27.96\% (HealthCareMagic, Qwen-14B), whereas the single-module variants still expose over 40\% of content. Averaged across all settings, the full method lowers the relative mean CRR to 0.51×, significantly outperforming RAGFort\textsubscript{InterOnly} (0.75×) and RAGFort\textsubscript{IntraOnly} (0.83×).
These results demonstrate that inter-class and intra-class protection are complementary, only their joint deployment can provide robust and comprehensive defense against knowledge extraction attacks.


\section{Limitation and Discussion}

\subsubsection{Experimental Scope and Systematic Evaluation.}
Due to page and resource constraints, we report only the primary experiments, including the dual-path defense comparison, security evaluation, system performance, and ablation study. Further studies can assess the robustness and practicality of RAGFort from more perspectives, such as parameter sensitivity and practical deployment efficiency.

\section{Conclusion}
In this paper, we address the problem of protecting proprietary knowledge bases in retrieval-augmented generation systems. Through systematic experiments, we demonstrate that effective protection requires jointly securing both intra-class and inter-class paths. Based on these insights, we propose RAGFort, a dual-path defense combining contrastive reindexing for inter-class isolation and cascade generation for intra-class robustness. Experimental results show that RAGFort significantly reduces knowledge base reconstruction while maintaining high answer quality and efficiency. To the best of our knowledge, RAGFort is the first systematic framework addressing dual-path protection in RAG systems, offering a practical solution for secure RAG deployments.

\newpage
\section*{Acknowledgements}
This work was supported by the Key Project of the National Natural Science Foundation of China under Grant no. 62536007, the Zhejiang Province Science Foundation under Grant no. LD24F020002 and the Zhejiang Province's 2025 “Leading Goose + X" Science and Technology Plan under Grant no. 2025C02034.

\bibliography{aaai2026}

\newpage
\appendix

\onecolumn
\section{\raggedright{Illustrative Case of the RAGFort Process}}
\label{appendix:case_study}
\begin{figure}[H]
\centering
\includegraphics[width=1\linewidth]{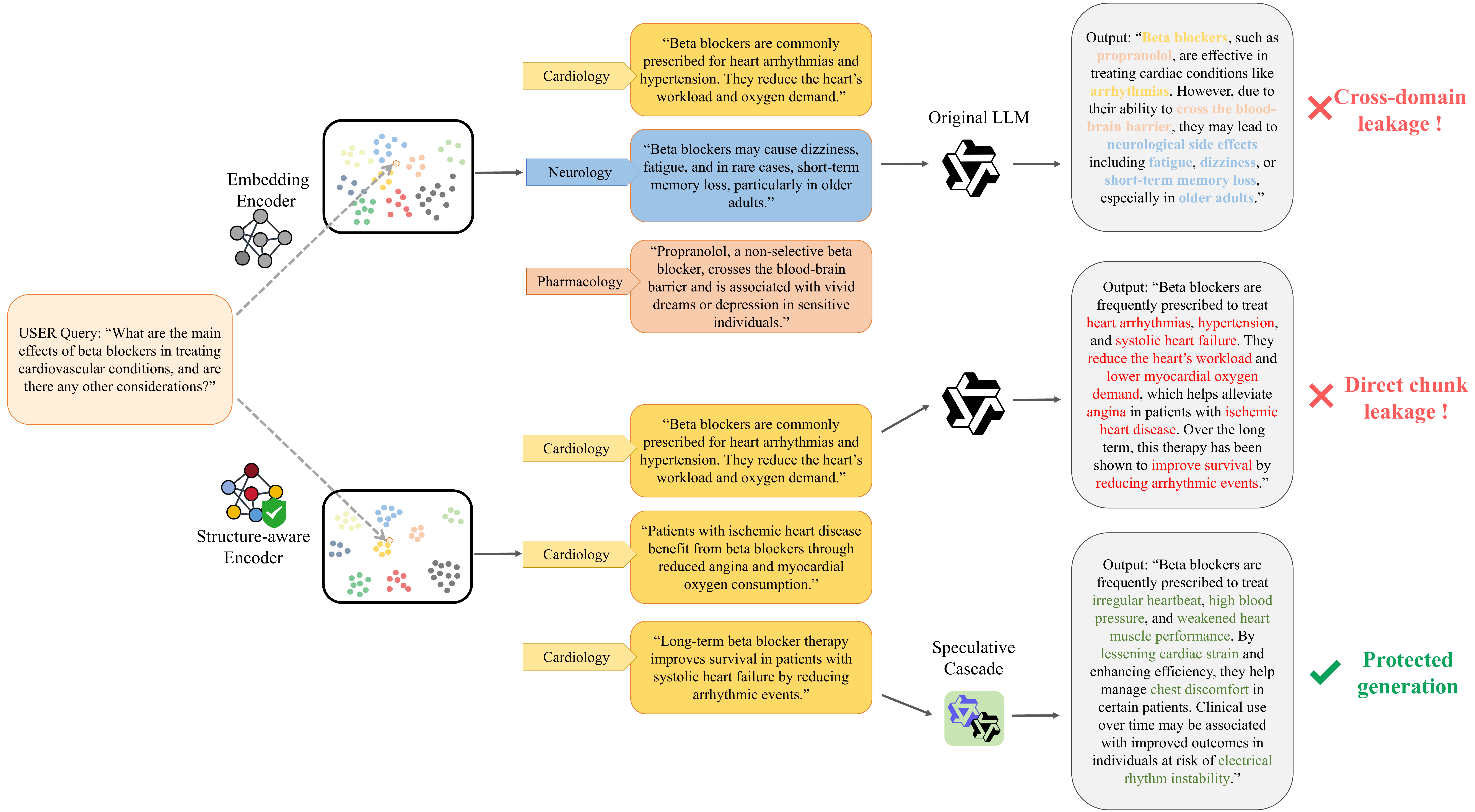} 
\caption{A representative example illustrating how RAGFort jointly applies inter-class and intra-class protection during inference.}
\label{fig:ragfort_case}
\end{figure}

\twocolumn
\section{Proofs of Lemmas}
\subsection{Proof of Lemma \ref{optimal}}
\textit{Proof.} Expanding the deferral risk in Equation~\eqref{eq:lagrangian}, we have:
\begin{equation}
\begin{aligned}
&L_{\text{spec}}(r; x_{<t},z)\\ 
=& r(x_{<t},z) \cdot ( \mathbb{E}_{x_t \sim \mathbb{P}(\cdot \mid x_{<t},z)} [\ell(x_t, p_t)] + \alpha \cdot D_{\text{TV}}(p_t, q_t) \\
-&\eta\cdot\frac{q_t(z)}{p_t(z)}- \mathbb{E}_{x_t \sim \mathbb{P}(\cdot \mid x_{<t},z)} [\ell(x_t, q_t)]) \\
+& \mathbb{E}_{x_t \sim \mathbb{P}(\cdot \mid x_{<t},z)} \left[\ell(x_t, q_t)\right]+\frac{q_t(z)}{p_t(z)}.
\end{aligned}
\end{equation}

This objective is minimized by a deferral rule $r : \mathcal{Y}^{t-1} \to \{0, 1\}$ that, for each prefix $x_{<t}$, chooses the value of $r(x_{<t})$ minimizing the total loss. Therefore, the optimal decision $r^*(x_{<t}) = 1$ if and only if the term within the parentheses is negative:
\begin{equation}
\begin{aligned}
&\mathbb{E}_{x_t \sim \mathbb{P}(\cdot \mid x_{<t},z)} \left[\ell(x_t, p_t)\right] + \alpha \cdot D_{\text{TV}}(p_t, q_t)\\
-&\eta\cdot\frac{q_t(z)}{p_t(z)} - \mathbb{E}_{x_t \sim \mathbb{P}(\cdot \mid x_{<t},z)} \left[\ell(x_t, q_t)\right] < 0,
\end{aligned}
\end{equation}
and $r^*(x_{<t}) = 0$ otherwise. Re-arranging the terms completes the proof.

\subsection{Proof of Lemma \ref{bound}}
For a fixed prefix $x_{<t}$ and $z$, we can write the deferral risk in Equation~\eqref{eq:lagrangian} as:
\begin{equation}
\begin{aligned}
&L_{\text{spec}}(r; x_{<t},z)\\ 
=& r(x_{<t},z) \cdot (\mathbb{E}_{x_t \sim \mathbb{P}(\cdot \mid x_{<t},z)} [\ell(x_t, p_t)] + \alpha \cdot D_{\text{TV}}(p_t, q_t)\\
-& \eta\cdot\frac{q_t(z)}{p_t(z)} - \mathbb{E}_{x_t \sim \mathbb{P}(\cdot \mid x_{<t},z)} [\ell(x_t, q_t)]) + Const,
\end{aligned}
\end{equation}
where $Const$ is a constant independent of the deferral rule. Let $r^* : \mathcal{Y}^{t-1} \to \{0, 1\}$ denote the optimal deferral rule. Then for any prefix $x_{<t}$ and $z$:
\begin{equation}
\begin{aligned}
&L_{\text{spec}}(\hat{r}_{\text{OPT}}; x_{<t},z) - L_{\text{spec}}(r^*; x_{<t},z) \\
=& (\hat{r}_{\text{OPT}}(x_{<t},z) - r^*(x_{<t},z)) \cdot ( \mathbb{E}_{x_t \sim \mathbb{P}} [\ell(x_t, p_t)] \\+& \alpha \cdot D_{\text{TV}}(p_t, q_t) -\eta\cdot\frac{q_t(z)}{p_t(z)} - \mathbb{E}_{x_t \sim \mathbb{P}} [\ell(x_t, q_t)] ).
\end{aligned}
\end{equation}

Adding and subtracting $\max_{y} q_t(y) - \max_{y} p_t(y)$ in the parentheses:
\begin{equation}
{\footnotesize
\begin{aligned}
&L_{\text{spec}}(\hat{r}_{\text{OPT}}; x_{<t},z) - L_{\text{spec}}(r^*; x_{<t},z)\\
=& (\hat{r}_{\text{OPT}}(x_{<t},z) - r^*(x_{<t},z)) \cdot( \max_{y} q_t(y) + \alpha \cdot D_{\text{TV}}(p_t, q_t)\\
-&\eta\cdot\frac{q_t(z)}{p_t(z)} - \max_{y} p_t(y)) \\
+& (\hat{r}_{\text{OPT}}(x_{<t},z) - r^*(x_{<t},z)) \cdot ( \mathbb{E}_{x_t \sim \mathbb{P}} [\ell(x_t, p_t)] - \\&\mathbb{E}_{x_t \sim \mathbb{P}} [\ell(x_t, q_t)] 
 - \max_{y} q_t(y) + \max_{y} p_t(y)).
\end{aligned}
}
\end{equation}

Using $\left| \hat{r}_{\text{OPT}}(x_{<t},z) - r^*(x_{<t},z) \right| \leq 1$, we upper-bound each term:
\begin{equation}
{\footnotesize
\begin{aligned}
\text{Term}_1 &= \max_{y} q_t(y) + \alpha \cdot D_{\text{TV}}(p_t, q_t)-\eta\cdot\frac{q_t(z)}{p_t(z)}- \max_{y} p_t(y), \\
\text{Term}_2 &= \mathbb{E}_{x_t \sim \mathbb{P}} [\ell(x_t, p_t)] - 1 + \max_{y} p_t(y), \\
\text{Term}_3 &= 1 - \max_{y} q_t(y) - \mathbb{E}_{x_t \sim \mathbb{P}} [\ell(x_t, q_t)].
\end{aligned}
}
\end{equation}

Hence, we have
\begin{equation}
\begin{aligned}
&L_{\text{spec}}(\hat{r}_{\text{OPT}}; x_{<t},z) - L_{\text{spec}}(r^*; x_{<t},z)\\ 
\leq &\text{Term}_1 + \text{Term}_2 + \text{Term}_3. \label{eq:lemma5-bound}
\end{aligned}
\end{equation}

We now bound each term. For Term$_1$, we consider two cases: (i) If $\max_{y} q_t(y) + \alpha \cdot D_{\text{TV}}(p_t, q_t) -\eta\cdot\frac{q_t(z)}{p_t(z)} - \max_{y} p_t(y) \leq 0$, then $\hat{r}_{\text{OPT}}(x_{<t},z) = 1$, and
{\footnotesize
\begin{equation}
\text{Term}_1 \leq \max_{y} p_t(y) + \alpha \cdot D_{\text{TV}}(p_t, q_t)-\eta\cdot\frac{\pi_t(z)}{q_t(z)} - \max_{y} q_t(y) \leq 0.
\end{equation}
}

(ii) Otherwise, $\hat{r}_{\text{OPT}}(x_{<t},z) = 0$, and Term$_1 = 0$ regardless of $r^*(x_{<t},z)$. Therefore, $\text{Term}_1\leq0$. Next, for Term$_2$, using $\ell = \ell_{0-1}$, we have:
\begin{equation}
{\footnotesize
\begin{aligned}
&\text{Term}_2\leq|\text{Term}_2|\\
=& |\mathbb{E}_{x_t \sim \mathbb{P}} \left[ \ell(x_t, p_t) \right] - 1 + \max_{y} p_t(y)| \\
=& |\sum_{x_t \in \mathcal{Y}} \mathbb{P}(x_t \mid x_{<t},z) \cdot \mathbb{I}\{x_t \neq \arg\max_{y} p_t(y)\} - 1 + \max_{y} p_t(y)| \\
=& |\max_{y} p_t(y) - \sum_{x_t \in \mathcal{Y}} \mathbb{P}(x_t \mid x_{<t},z) \cdot \mathbb{I}\{x_t = \arg\max_{y} p_t(y)\}| \\
=& \left| p_t(y^*) - \mathbb{P}(y^* \mid x_{<t},z) \right| \leq \max_{y} \left| p_t(y) - \mathbb{P}(y \mid x_{<t},z) \right|,
\end{aligned}
}
\end{equation}
where $y^* = \arg\max_{y} p_t(y)$.

Similarly, for Term$_3$:
\begin{equation}
\begin{aligned}
\text{Term}_3 \leq \max_{y} \left| q_t(y) - \mathbb{P}(y \mid x_{<t},z) \right|.
\end{aligned}
\end{equation}

Substituting the bounds for Term$_1$–Term$_3$ into Equation~\eqref{eq:lemma5-bound} completes the proof.

\end{document}